\documentclass[journal]{IEEEtran}
\usepackage{xcolor}
\usepackage{booktabs}
\usepackage{epsfig}
\usepackage{graphicx}
\usepackage{amsmath}

\usepackage{amssymb}
\usepackage{amsmath}
\usepackage{verbatim}

\usepackage{url}
\usepackage{tabularx}
\usepackage{multirow}
\usepackage{subfigure}
\usepackage{amsmath,dsfont}
\usepackage{array}
\usepackage{enumitem}

\usepackage{pifont}
\usepackage{threeparttable}
\usepackage{bbding}
\usepackage{cleveref}
\usepackage{dashrule}

\usepackage{verbatim}

\usepackage{colortbl}
\definecolor{mygray}{gray}{.75}

\usepackage{xcolor}

\usepackage{pifont}

\usepackage[normalem]{ulem}

\usepackage{makecell}
\usepackage{multirow}
\usepackage[linesnumbered,ruled,vlined]{algorithm2e}
\usepackage{algorithmic}

\begin{document}
%

\title{RSRNav: Reasoning Spatial Relationship for Image-Goal Navigation}
%
%
%

\author{Zheng~Qin,
        Le~Wang$^{*}$,~\IEEEmembership{Senior Member,~IEEE},
        Yabing Wang,
        Sanping Zhou,~\IEEEmembership{Member,~IEEE,}
        Gang Hua,~\IEEEmembership{Fellow,~IEEE,}
        Wei Tang,~\IEEEmembership{Member,~IEEE}

	\thanks{

This work was supported in part by National Science and Technology Major Project under Grant 2024YFB4708100, National Natural Science Foundation of China under Grants 62088102, 12326608 and 62106192, Natural Science Foundation of Shaanxi Province under Grant 2022JC-41, Key R\&D Program of  Shaanxi Province under Grant 2024PT-ZCK-80, and Fundamental Research Funds for the Central Universities under Grant XTR042021005. \textit{(Corresponding author: Le Wang.)}}%
	\thanks{Zheng Qin, Le Wang, Yabing Wang and Sanping Zhou are with the National Key Laboratory of Human-Machine Hybrid Augmented Intelligence, National Engineering Research Center for Visual Information and Applications, and Institute of Artificial Intelligence and Robotics, Xi'an Jiaotong University, Xi'an, Shaanxi 710049, China. (e-mail: qinzheng@stu.xjtu.edu.cn; wyb7wyb7@163.com; \{lewang, spzhou\}@mail.xjtu.edu.cn)}
	\thanks{Gang Hua is with the Multimodal Experiences Lab, Dolby Laboratory, Bellevue, WA 98004, USA. (e-mail: ganghua@gmail.com)}
        \thanks{Wei Tang is with the Department of Computer Science, University of Illinois, Chicago, IL 60607, USA(e-mail: tangw@uic.edu)}}

\markboth{IEEE Transactions on Circuits and Systems for Video Technology,~Vol.~x, No.~x}
{Shell \MakeLowercase{\textit{et al.}}: Bare Demo of IEEEtran.cls for IEEE Journals}
%



\maketitle

\begin{abstract}
 Recent image-goal navigation (ImageNav) methods learn a perception-action policy by separately capturing semantic features of the goal and egocentric images, then passing them to a policy network. However, challenges remain: (1) Semantic features often fail to provide accurate directional information, leading to superfluous actions, and (2) performance drops significantly when viewpoint inconsistencies arise between training and application.
To address these challenges, we propose RSRNav, a simple yet effective method that reasons spatial relationships between the goal and current observations as navigation guidance. 
Specifically, we model the spatial relationship by constructing correlations between the goal and current observations, which are then passed to the policy network for action prediction. These correlations are progressively refined using fine-grained cross-correlation and direction-aware correlation for more precise navigation.
Extensive evaluation of RSRNav on three benchmark datasets demonstrates superior navigation performance, particularly in the "user-matched goal" setting, highlighting its potential for real-world applications.
\end{abstract}


%
\IEEEpeerreviewmaketitle

\section{Introduction}
Visual navigation~\cite{he2024memory, fang2021visual,he2025navcomposer,cao2025erd} emerges as a pivotal research area within embodied artificial intelligence, requiring an agent to navigate unfamiliar environments to reach specific goals. Although multiple information carriers can represent goals, images offer a more accurate and detailed description for specifying the goal's location. Our research focuses on the image-goal navigation task (ImageNav), a field brimming with diverse potential applications, such as household robots, augmented reality systems, and assistance for individuals with visual impairments.

\begin{figure*}[!t]
	\centering
	\includegraphics[width=0.8\linewidth]{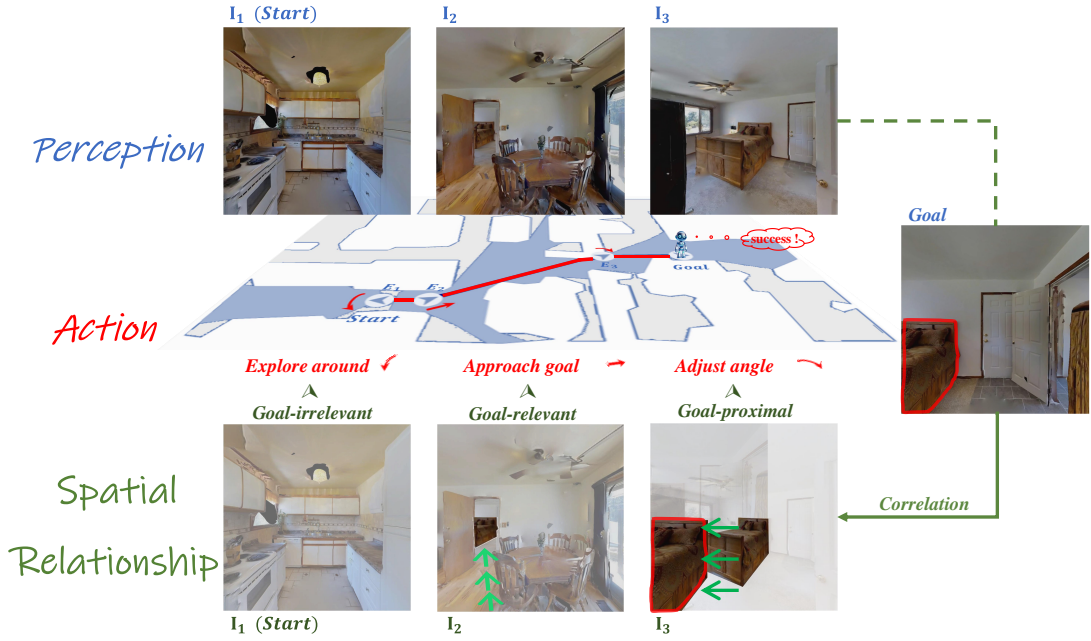}
	\caption{Illustration of human-intuitive navigation, with \textcolor{red}{action} decisions guided by constantly reasoning  \textcolor[RGB]{84,130,53}{spatial relationship} between the current observation $\mathbf{I}_1, \mathbf{I}_2, \mathbf{I}_3$ and the goal~(\textcolor[RGB]{60,107,194}{perception}). Goal-irrelevant $\mathbf{I}_1$: the current observation is irrelevant to the goal; we should either turn around or explore a new area. Goal-relevant $\mathbf{I}_2$: the element relevant to the goal is encountered, \emph{i.e.}, the bed; we stop exploring and approach the goal under the bed's spatial guidance. Goal-proximal $\mathbf{I}_3$: We are close to the goal; we precisely adjust the angle based on the relative positional relationship of the goal in the current observation. }
	\label{fig:intro}
\end{figure*}

Existing ImageNav methods generally follow the modular paradigm ~\cite{chaplot2020learning} or the end-to-end paradigm~\cite{yadav2023offline}. 
In modular methods, every module is optimized independently, with the system's overall performance being limited by the capabilities of its individual modules. Additionally, it requires extra information beyond its egocentric capture, such as depth information and GPS position. 
End-to-end methods have attracted increasing attention due to the streamlined process and global optimization, enabling direct learning from data. The core of these end-to-end methods lies in training a perception-action policy, typically facilitated by a visual encoder for perception. Encoders are trained along with the policy network, specifically dedicated to furnishing more effective information for action prediction. 
For example, the classic framework~\cite{al2022zero} employs two separate encoders to extract semantic embeddings from goal and egocentric images independently, which are then concatenated to serve as cues for action prediction.
Recent efforts have focused on enhancing the encoder's ability for feature extraction, such as offline pretraining of visual representations with self-supervised learning~\cite{yadav2023offline}; utilization of powerful encoder structures~\cite{yadav2023ovrl}; enhancement of image embeddings using the CLIP model~\cite{majumdar2022zson}; introduction of memory modules~\cite{kwon2021visual,li2024memonav}, \emph{etc}. Despite the remarkable advances in extracting semantic embeddings, there are still some challenges regarding the efficiency and robustness of navigation for these feature-dependent methods.

First, semantic feature vectors cannot provide the agent with bearing information about the goal within the scene. This issue may cause superfluous actions in the attempt to navigate and properly orient toward the goal, which would affect navigation efficiency. Second, in real-world applications, users would not capture goal images from the same viewpoint as the agent, which is inconsistent between training and application. Relying on
semantic information generated by unseen viewpoint images
can bias the policy network in predicting action, leading to a substantial decrease in navigation performance.
Both of these challenges stem from the limitations of semantic features within this framework.

More human-intuitive navigation is achieved efficiently by reasoning about the spatial relationship between the current environment and the target position. This spatial relationship can provide rich orientation information to guide navigation actions. As shown in~\Cref{fig:intro}, the agent explores new areas when the observation is goal-irrelevant~($\mathbf{I}_1$); approaches the goal when the observation is goal-relevant~($\mathbf{I}_2$); and precisely adjusts the angle when the observation is goal-proximal~($\mathbf{I}_3$). 
Furthermore, spatial relationship between images are less affected by inconsistent viewpoints because they focus on how images relate to each other rather than on the detailed semantics information. Inspired by these, we aim to overcome the above challenges by reasoning spatial relationship and using it to guide navigation actions.

In this paper, we model the spatial relationship between the goal and the observations by constructing correlations between the features extracted from the goal and observation images in the perception step and pass them information to the policy network, \emph{i.e.}, training a perception-relationship-action navigation policy. We first devise an extremely simple version called ``Navigation Via Minimalist Relationship'' to explore the importance of spatial relationship. It simply passes two correlation scores to the policy network yet surprisingly outperforms the baseline. We then evolve the relationship modeling to the ``Navigation Via Dense Relationship'' version, which models richer and more detailed relationship by calculating cross-correlation with finer granularity. The significant improvement further confirms the critical role of spatial relationship in ImageNav. Finally, we reinforce the model by introducing the correlation that contains directional information, named ``Navigation Via Direction-aware Relationship''. It is implemented by constructing a multi-scale correlation matrix and lookup with a directional-aware operator, making the agent's navigation more efficient and enabling more precise angle adjustments. Extensive experiments on three benchmark datasets (Gibson, MP3D, and HM3D) show that our proposed RSRNav performs well in terms of navigation efficiency~(SPL) and significantly outperforms previous state-of-the-art methods in all metrics under the ``user-matched goal'' setting.


The main contribution of this work can be summarized as follows:
\begin{itemize}
    \item We propose a novel method for efficient and robust ImageNav by constantly reasoning about the spatial relationship between the goal and current observations as navigation guidance during the navigation process.
    \item We model the spatial relationship by calculating the correlations between features of the goal image and the current observation and pass it to the policy network, \emph{i.e.}, training a perception-relationship-action navigation policy. We gradually enhance the relationship modeling by using more fine-grained cross-correlation and introducing direction-aware correlation to provide precise direction information for navigation.
    \item Extensive experiments demonstrate the superior performance of our proposed RSRNav in terms of navigation efficiency~(SPL). It significantly outperforms previous state-of-the-art methods across all metrics under the ``user-matched goal'' setting, showing potential for real-world applications.
\end{itemize}

The remainder of the paper is organized as follows.
We briefly review the related work in Section~\ref{sec:relatedWork}.
Subsequently, we present the technical details of the proposed method in Section~\ref{sec:method}.
The experimental results are presented in Section~\ref{sec:experiment}.
Finally, we conclude this paper in Section~\ref{sec:conclusion}.

\section{Related Work}\label{sec:relatedWork}
\subsection{Modular Methods}
Modular methods for navigation break down the high-level tasks into a series of independent sub-tasks or modules responsible for handling a specific function within the navigation process, such as classic perception~\cite{yamauchi1997frontier}, localization, map construction~(SLAM~\cite{ lin2025slam2}), and path planning. Some of these modules require no extra training and fine-tuning~\cite{krantz2023navigating}, and some need additional data to support the training~\cite{ ramrakhya2022habitat, wasserman2023last}. 
As an example, \cite{krantz2023navigating} tackles various sub-tasks, including exploration, goal re-identification, goal localization, and local navigation. This work uses feature-matching methods to re-identify the goal in egocentric vision and then project the matched features onto a map for localization. All of these tasks are addressed using standard components, without requiring any finetuning.
Some methods build semantic maps~\cite{chaplot2020neural, hahn2021no} to help agents with better localization and planning. For instance, \cite{chaplot2020neural} introduces topological models of space that incorporate semantic information, enabling coarse geometric reasoning and improving resilience to actuation errors. By storing semantic features at nodes, the system can exploit statistical patterns, facilitating efficient exploration in unfamiliar environments. 
The recent work~\cite{johnson2024feudal} uses Superglue~\cite{sarlin2020superglue} to provide directional guidance by matching the current and goal images' keypoints.
Some methods discard map construction and simply use modern vision tasks, such as object detection or segmentation~\cite{ramrakhya2022habitat, maksymets2021thda}. 

\subsection{End-to-end Methods}
In contrast to module methods, end-to-end methods allow good performance without using semantic mapping, object detection, segmentation, \emph{etc}. End-to-end methods~\cite{majumdar2022zson,yadav2023offline} directly map inputs~(from the sensors) to agent’s action through learning a navigation policy in an end-to-end manner using reinforcement~\cite{zhu2017target} or imitation learning~\cite{ding2019goal}. 

The core of end-to-end methods is to train a perception-to-action policy. 
CRL~\cite{du2021curious} simultaneously trains a reinforcement learning policy and a visual representation model. The policy is designed to maximize the errors in the representation model, which encourages the agent to explore its environment. As the policy generates increasingly challenging data, the representation model becomes progressively more refined and robust.
ZER \cite{al2022zero} introduces a zero-shot transfer learning framework that incorporates an innovative reward mechanism for its semantic search policy. This approach allows the model to effectively transfer knowledge gained from a source task to multiple target tasks (such as ObjectNav, RoomNav, and ViewNav), which involve diverse goal modalities like images, sketches, audio, and labels. Additionally, this model supports zero-shot experience learning, enabling it to address target tasks without requiring any task-specific interactive training.
Some other efforts and advances have focused on enhancing the agent's perceptual capabilities, \emph{i.e.}, how to encode the observation's features and the goal better. For example, ZSON~\cite{majumdar2022zson} uses the CLIP model to improve image embeddings, which has been widely used in the multimodal field~\cite{wang2024dual,dong2022reading,wang2025denoising,liu2025up}.  
OVRL~\cite{yadav2023offline} introduces a two-step approach: (1) offline pretraining of visual features through self-supervised learning (SSL) on a large dataset of pre-rendered indoor environment images, and (2) online fine-tuning of visuomotor representations for specific tasks, utilizing image augmentations and extended training periods.
After that, OVRL-V2~\cite{yadav2023ovrl} introduces a stronger model, pre-training vision transformers~(ViT). 
FGPrompt~\cite{sun2024fgprompt} proposes fusion strategies to extract the features of observation based on the goal prompt. And the memory module has been introduced to address the long-horizon planning~\cite{he2024memory}. 
In distinction to previous works focusing on extracting better and more efficient features, we propose to model the spatial relationship between current observation and the goal and train a perception-relationship-action policy.




\section{Method}\label{sec:method}

\subsection{Task Definition and Overview of RSRNav}
\label{overview}
\noindent{\bf Task Definition.} 
In an image-goal navigation episode, the target of the agent is to reach the goal position, denoted as $\mathbf{p}_g$, guided by a goal image $\mathbf{I}_g$ that was taken at $\mathbf{p}_g$. Specifically, the agent is assigned to start from a randomly chosen position $\mathbf{p}_0$ within an unexplored scene, using an egocentric camera to guide its navigation. At each time step $t$, the agent receives an egocentric RGB image~$\mathbf{I}_t$  from its body-mounted RGB sensor and predicts its following action from \{move forward, turn left, turn right, stop\}. Navigation is deemed successful if the agent stops within a specified distance of $\mathbf{p}_g$.

Furthermore, the parameters of the goal camera are the same as those of the agent's egocentric camera, leading to inconsistencies between training and applications due to the different perspectives of the agent and the user, such as heights, pose, visual angles, \emph{etc}. We use two settings, ``agent-matched goal'' and ``user-matched goal'' during the evaluation episodes following~\cite{ sun2024fgprompt}.

\begin{figure*}[!t]
	\centering
	\includegraphics[width=1\linewidth]{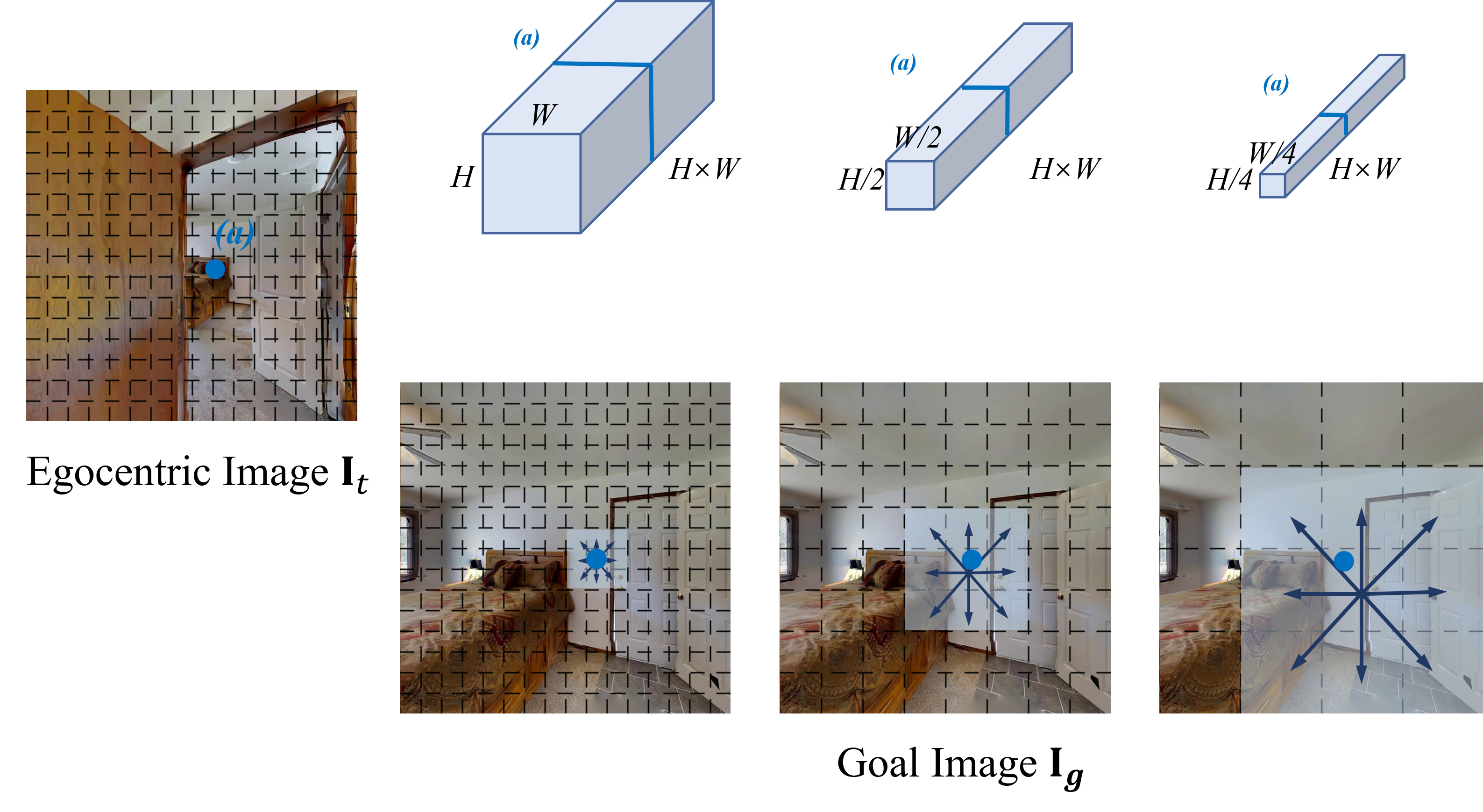}
	\caption{Overview of our RSRNav. ``Minimalist Relationship'' constructs sparse correlations to initially explore the importance of spatial relationship  in ImageNav. ``Dense Relationship'' uses cross-correlation with finer granularity to establish richer and more detailed relationship. ``Direction-aware Relationship'' introduces direction information to support the agent in moving more efficiently and adjusting the angle more precisely.}
	\label{fig:method}
\end{figure*}

\noindent{\bf Overview.} 
The core of our RSRNav is to train a perception-relationship-action navigation policy. 
At each time step $t$, the agent receives an egocentric RGB image $\mathbf{I}_t$, and conducts feature extraction along with the goal image $\mathbf{I}_g$ in the \textbf{perception} step. Subsequently, we reason the \textbf{relationship} by calculating the correlations between these features, generating the correlation cue $\mathbf{C}^\text{cue}$ that is fed into the policy network for \textbf{action} prediction.

We gradually reinforce the correlation calculation with three versions and eventually develop a powerful direction-aware correlation for reasoning spatial relationship.
First, we explore the importance of spatial relationship in the navigation task, \emph{i.e.}, is the spatial relationship information critical for cueing action prediction among a large pile of information that passes to the policy network. For this, we design the ``Navigation Via \textbf{Minimalist Realtionship}'' version, relying only on two correlation scores to predict actions. 
With the exciting discovery that it outperforms the baseline with such sparse correlation, we continue to develop the potential of this relationship-based method. 
We further enhance the spatial relationship modeling and evolve it to the ``Navigation Via \textbf{Dense Realtionship}'' version. 
Finally, we introduce ``Navigation Via \textbf{Direction-aware Realtionship}'' version to furnish more direction information to support the agent in moving more efficiently and adjusting the angle more precisely.

\subsection{Perception}
\label{Perception}
Unlike prior methods~\cite{majumdar2022zson, yadav2023offline} that rely on complex, pre-trained encoder networks, our method employs a simple ResNet-9 without any pre-training. Furthermore, we utilize a weight-sharing network for both rather than using separate networks to extract features from egocentric and goal images. At time step $t$, we employ a weight-sharing ResNet-9 to encode both the goal image $\mathbf{I}_g$ and the current egocentric image $\mathbf{I}_t$ into features~(feature vectors or feature maps), as shown in~\Cref{fig:method}.

\subsection{Reasoning Spatial Relationship}
\label{Correlation}
\subsubsection{Minimalist Realtionship}~
\label{Minimalist Correlation}

\noindent In the perception step, $\mathbf{I}_g$ and $\mathbf{I}_t$ 
are encoded into sparse feature vectors denoted as $\mathbf{V}_{g}$, $\mathbf{V}_{t} \in \mathbb{R}^{2 \times  D}$, 
where $D$ is the feature dimension. The receptive fields of the two feature vectors approximate the left and right halves of the picture, respectively. Then we calculate the correlation of the left and right halves of $\mathbf{I}_t$ and $\mathbf{I}_g$ by dot product on the corresponding feature vectors. The dot product results are normalized as feature similarity and form correlation cue $\mathbf{C}^\text{cue} \in \mathbb{R}^{2}$, which contains the relation of the left and right counterparts of  $\mathbf{I}_t$ and $\mathbf{I}_g$. We use it as a basis for action prediction.

\noindent{\bf Discussion.} 
In this part, we present a minimalist relationship-based method, where only two scores are passed to the policy network for both training and inferring. Compared to the baseline, where complete semantic information is passed to the policy network, our correlation cue contains much less information but still outperforms the baseline. This reflects two aspects: Although semantic features contain a lot of information, some of them are not critical and instead increase the burden on training the policy network; the spatial relationship plays a key role in navigation and deserves more attention.

\subsubsection{Dense Realtionship}~
\label{Dense Correlation}

\noindent To establish richer and more detailed relationship information, we reinforce ``Minimalist Relationship'' by utilizing cross-correlation with finer granularity. Instead of encoding images into sparse feature vectors with large receptive fields, we encode $\mathbf{I}_t$ and $\mathbf{I}_g$ into dense feature maps, denoted as $\mathbf{F}_{g}$, $\mathbf{F}_{t} \in \mathbb{R}^{H \times W \times  D}$, where $H, W$ are respectively 1/32 of the raw image size, and $D$ is the feature dimension.

After obtaining the feature maps $\mathbf{F}_{g}$ and $\mathbf{F}_{t}$, we compute the cross-correlation by a cross-correlation layer. This layer is not limited to corresponding feature points, but contains a dot product operation for all feature point pairs between two feature maps and constructs a correlation matrix $\mathbf{C}$ as follows:
\begin{equation}
\begin{aligned}
	&\mathbf{C}(i, j)=\left\langle\mathbf{F}_{g}(i),\mathbf{F}_{t}(j)\right\rangle \in \mathbb{R}^{H \times W \times H \times W},
\end{aligned}
\end{equation}
where $i, j$ indicate the index of feature map $\mathbf{F}_{g}$ and $\mathbf{F}_{t}$. The element $c_{ij}$ in $\mathbf{C}$ represents the similarity between the $i$-th feature point in $\mathbf{F}_{g}$ and the $j$-th feature point in $\mathbf{F}_{t}$.
$\mathbf{C}$ contains the dense global similarity between the current observation and the goal.
Then we flatten the 4D matrix to form the correlation cue $\mathbf{C}^\text{cue}$ as follows:
\begin{equation}
\begin{aligned}
	\mathbf{C}^\text{cue} &= \mathtt{flatten}( \mathbf{C} )  \in \mathbb{R}^{1 \times H^2  W^2}.
\end{aligned}
\end{equation}

\begin{figure}[t]
  \centering
\includegraphics[width=0.8\linewidth]{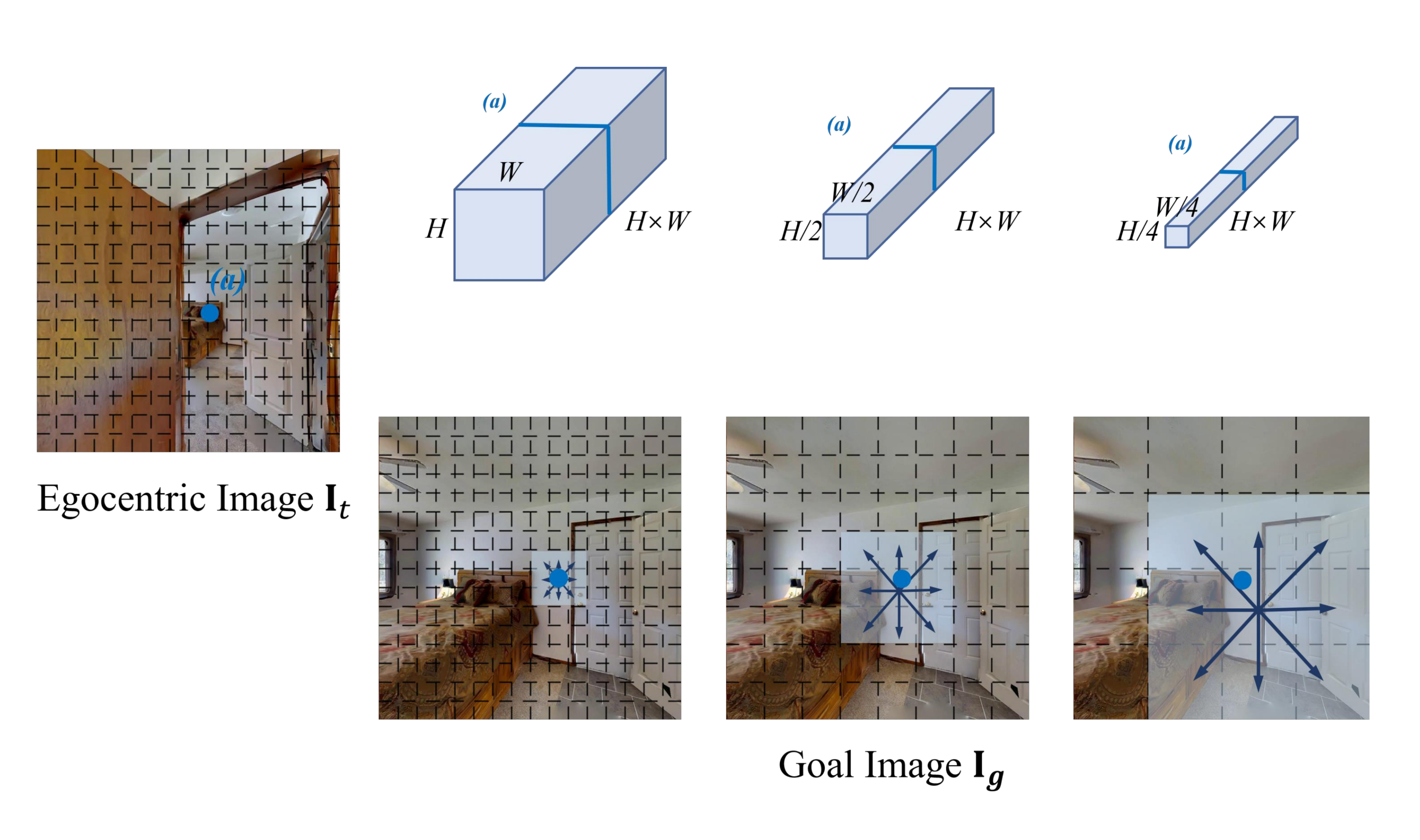}
  \caption{Graphical representation of the correlation pyramid and center-based direction-aware searching region. The black grids represent the approximate receptive fields of the feature points of $\mathbf{F}_{t}$ and the regions of $\mathbf{F}_{g}$ at different scales on the raw image in the correlation pyramid. The light blue areas depict the searching region template for each feature point in $\mathbf{F}_{t}$, covering the surrounding area in all directions.}
  \label{fig:method2}
\end{figure}

\subsubsection{Direction-aware Realtionship}~
\label{Direction-aware Correlation}

\noindent To enable the agent's navigation to be more efficient and angle adjustments to be more precise, it is necessary to provide the policy network with effective direction information from the current observation to the target. 

On the basis of ``Dense Relationship'', we downsample the correlation matrix $\mathbf{C}$, obtaining the correlation pyramid $\{\mathbf{C}_s\}_{s=0}^S$ by pooling the last two dimensions as follows:
\begin{equation}
\begin{aligned}
	\mathbf{C}_s = \mathtt{pooling}(\mathbf{C}_{\emph{s}-1}), \mathbf{C}_\emph{s} \in \mathbb{R}^{\emph{H} \times \emph{W} \times \emph{H}/2^\emph{s} \times \emph{W}/2^\emph{s}}, \\ \\
\end{aligned}
\end{equation}
where $s$ is the layer of pyramid and $\mathbf{C}_0$ is initialized as $\mathbf{C}$. This set of matrices provides the similarity between each feature point in $\mathbf{F}_{t}$ and every region in $\mathbf{F}_{g}$ at different scales.
The black grids that we illustrate in~\Cref{fig:method2} represent the approximate receptive fields of the feature points in $\mathbf{F}_{t}$ and the regions in $\mathbf{F}_{g}$ at different scales on the raw image.

To consider the precise direction and reduce the significant number of invalid correlations, we set a center-based direction-aware searching region to lookup on $\{\mathbf{C}_s\}_{s=0}^S$ as follows:
\begin{equation}
\begin{aligned}
        &\mathbf{O}_s = \mathtt{Lookup}(\mathbf{C}_\emph{s}, \mathcal{P}\left(\mathbf{x}\right)_\emph{s}), \\
	&\mathcal{P}\left(\mathbf{x}\right)_s=\left\{\mathbf{x}/2^s+\mathbf{r}  : \|\mathbf{r}\| \leq 1\right\}, \\
\end{aligned}
\end{equation}
$\mathcal{P}\left(\mathbf{x}\right)_\emph{s}$ defines a square region of size $3 \times 3$ for each position $\mathbf{x}$ in $\mathbf{F}_{t}$. Cropping this region in the channel corresponding to $\mathbf{x}$ in the correlation matric $\mathbf{C}_s$, followed by reshaping, results in a $3^2$-dimensional correlation vector. In~\Cref{fig:method2}, the correlation vectors for each scale are depicted within the light blue regions of the goal image. Collecting these vectors for all $\mathbf{x}$ in $\textbf{F}_t$ leads to $\mathbf{O}_s\in \mathbb{R}^{H\times W\times 3^2}$.

We then concatenate different scales' $\mathbf{O}_s$ and form $\mathbf{O} \in \mathbb{R}^{H \times W \times ((S+1)\times3^2)}$, in which each feature point represent a multi-scale correlation vector. Following this, a series of convolutional layers are implemented on $\mathbf{O}$ to further fuse the correlation features, which are finally flattened as follows:
\begin{equation}
\begin{aligned}
	\mathbf{C}^\text{cue} &= \mathtt{flatten}(\mathtt{conv}(\mathtt{concat}(\left\{\mathbf{O}_\emph{s}\right\}_{\emph{s}=0}^\emph{S}))) ,
\end{aligned}
\end{equation}

\subsection{Learning Navigation Policy}
\label{policy}
We train the navigation policy $\pi$ using reinforcement learning. As shown in \Cref{fig:method}, the agent obtains its current correlation cue $\mathbf{C}^\text{cue}$ and passes it to the policy $\pi$. Following previous methods~\cite{majumdar2022zson,sun2024fgprompt,al2022zero}, the policy further encodes this information along with the history so far to produce a state embedding $\mathbf{s}_{t}$. An actor-critic network leverages $\mathbf{s}_{t}$ to predict state value $\mathbf{c}_{t}$ and the agent’s next action $\mathbf{a}_{t}$. The model is trained end-to-end using PPO~\cite{schulman2017proximal} with the reward in~\cite{al2022zero}.

In details, we train the model with the reward received for performing a particular action in the environment. This reward signal instructs the agent to learn what behaviors are good (\textit{i.e.}, receive a positive reward) or what behaviors are bad (\textit{i.e.}, receive a negative reward or a lower positive reward). Following~\cite{al2022zero}, the reward consists of two sub-reward functions. The first is calculated at each time step in the episode as follows:
\begin{equation}
\begin{aligned}
r_t=r_d\left(d_t, d_{t-1}\right)+\left[d_t \leq d_s\right] r_\alpha\left(\alpha_t, \alpha_{t-1}\right)-\gamma,
\end{aligned}
\end{equation}
where $d_t$ is the distance to the goal at time step $t$ and $\alpha_t$ is the angle in radians to the goal view. $r_d$ is the reduced distance to the goal from the current position relative to the previous one, and $r_\alpha$ is the reduced angle in radians to the goal view from the current view relative to the previous one. [·] is the indicator function. $\gamma$~=~0.01 is to encourage reaching the goal in fewer steps because the more steps are taken, the more penalties $\gamma$'s accumulation brings.

The second reward only exists at the end of an episode, as follows:
\begin{equation}
\begin{aligned}
r_{\text{final}}=5 \times\left(\left[d_t \leq d_s\right]+\left[d_t \leq d_s \text { and } \alpha_t \leq \alpha_s\right]\right),
\end{aligned}
\end{equation}
When the agent gives a stop command, if the agent stops within a of the specific distance $d_s$ from the goal, it gets a reward of 5. In addition to this, when the angle between the agent's pose and the goal view is small, an additional reward of 5 is reaped.


\begin{table*}[tb]
\caption{Comparison with the state-of-the-art methods on Gibson. The two settings, ``agent-matched goal'' and ``user-matched goal'', represent the goal camera reflecting the task executor and task publisher, respectively. The best results for each metric are highlighted in bolded. We report the ImageNav results averaged over three random seeds.}
\label{tab:table1}
\centering
\resizebox{0.8\linewidth}{!}{
\begin{tabular}{lcccccc}
\hline
\textbf{Method}                     & \textbf{Venue} & \textbf{Backbone}    & \textbf{Pretrain}    & \textbf{Sensor(s)}   & \textbf{SPL}         & \textbf{SR}          \\ \hline
\emph{\textbf{agent-matched goal}} &                & \multicolumn{1}{l}{} & \multicolumn{1}{l}{} & \multicolumn{1}{l}{} & \multicolumn{1}{l}{} & \multicolumn{1}{l}{} \\
Act-Neur-SLAM~\cite{chaplot2020learning}                       &     ICLR20           & ResNet9              & \ding{55}                   & 1RGB+Pose             & 23.0\%               & 35.0\%               \\
NTS~\cite{chaplot2020neural}                                 &     CVPR20           & ResNet9              & \ding{55}                   & 1RGBD+Pose            & 43.0\%               & 63.0\%               \\ 
Mem-Aug~\cite{mezghan2022memory}                             &    IROS22            & ResNet18             & \ding{55}                   & 4 RGB                & 56.0\%               & 69.0\%               \\
VGM~\cite{kwon2021visual}                                 &     ICCV23           & ResNet18             & \ding{55}                   & 4 RGB                & 64.0\%               & 76.0\%               \\
TSGM~\cite{kim2023topological}                        &     CoRL23           & ResNet18             & \ding{55}                   & 4 RGB                & 67.2\%               & 81.1\%               \\ \hline
ZER~\cite{al2022zero}               &    CVPR22            & ResNet9              & \ding{55}                   & 1RGB                  & 21.6\%               & 29.2\%               \\
ZSON~\cite{majumdar2022zson}        &    NIPS22            & ResNet50             & \checkmark                  & 1RGB                  & 28.0\%               & 36.9\%               \\
OVRL~\cite{yadav2023offline}        &    ICLR22            & ResNet50             & \checkmark                  & 1RGB                  & 27.0\%               & 54.2\%               \\
OVRL-V2~\cite{yadav2023ovrl}        &  arXiv23              & ViT             & \checkmark                 & 1RGB                  & 58.7\%               & 82.0\%               \\
FGPrompt-MF~\cite{sun2024fgprompt}     &   NIPS23              & ResNet9              & \ding{55}                   & 1RGB                  & 62.1\%               & 90.7\%      \\
FGPrompt-EF~\cite{sun2024fgprompt}     &   NIPS23              & ResNet9              & \ding{55}                   & 1RGB                  & 66.5\%               & 90.4\%      \\
\rowcolor[HTML]{DAE8FC} RSRNav~(ours)        &                & ResNet9              & \ding{55}                   & 1RGB                  & \textbf{69.5\%}      & \textbf{91.1}\%               \\ \hline 
\emph{\textbf{user-matched goal}}    &       -         & \multicolumn{1}{l}{} & \multicolumn{1}{l}{} & \multicolumn{1}{l}{} & \multicolumn{1}{l}{} & \multicolumn{1}{l}{} \\
ZER\cite{al2022zero}                            &     CVPR22          & ResNet9              & \ding{55}                   & 1RGB                  & 12.8\%               & 15.7\%               \\
FGPrompt-MF\cite{sun2024fgprompt}                            &      NIPS23          & ResNet9              & \ding{55}                   & 1RGB                  & 40.3\%               & 69.2\%               \\
FGPrompt-EF\cite{sun2024fgprompt}                            &      NIPS23          & ResNet9              & \ding{55}                   & 1RGB                  & 46.8\%               & 76.0\%               \\
\rowcolor[HTML]{DAE8FC} RSRNav~(ours)          &        -        & ResNet9              & \ding{55}                   & 1RGB                  & \textbf{56.6\%}      & \textbf{83.2\% }      \\ \hline  
\end{tabular}}
\end{table*}

\section{Experiments}
\label{sec:experiment}
\subsection{Setting}
\label{Setting}
\noindent{\bf Task Setup and Evaluation Metrics.}\hspace{0.3cm}
We adopt the image-goal navigation task defined in~\Cref{overview}. Navigation is considered successful when the agent stops within 1 m of $\mathbf{p}_g$, within the maximum of 500 steps per episode. Following~\cite{al2022zero, yadav2023offline, sun2024fgprompt}, the action space of the agent is \{move forward 0.25m, turn left $30^\circ$, turn right $30^\circ$, stop\}. The agent uses only an egocentric RGB camera of 128 × 128 resolution and a $90^\circ$ FoV sensor. To quantitatively evaluate navigation performance, we employ the success rate~(SR) and Success weighted by Path Length~(SPL), which takes into account navigation's path efficiency~\cite{anderson2018evaluation}. 

\noindent{\bf Goal Camera Setting.}\hspace{0.3cm}
We provide evaluation results under two goal camera settings, as described in~\Cref{overview}. Under the setting of ``agent-matched goal'', the parameters of the goal camera are consistent with the agent's egocentric camera. For the ``user-matched goal'' setting, the goal camera parameters are more closely aligned with the human user, which is inconsistent with the agent's egocentric camera. We sample the height of goal camera from $\mathcal{U}(0.8m, 1.5m)$, pitch delta from $\mathcal{U}(-5^\circ, 5^\circ)$, and the HFOV from $\mathcal{U}(60^\circ, 120^\circ)$ following~\cite{sun2024fgprompt, krantz2022instance}.

\noindent{\bf Datasets.}\hspace{0.3cm}
We employ the Habitat simulator~\cite{savva2019habitat} and the Gibson environments~\cite{xia2018gibson} for training our model, utilizing the dataset split provided in~\cite{mezghan2022memory}. The training set consists of 9,000 episodes from each of the 72 scenes. The test set comprises 4,200 episodes sampled from 14 unseen scenes. And we test our Gibson-trained models on MP3D~\cite{chang2017matterport3d} and HM3D~\cite{ramakrishnan2021habitat} to evaluate the performance of domain generation.

\noindent{\bf Implementation Details.}\hspace{0.3cm}
We implemented our RSRNav in PyTorch~\cite{paszke2019pytorch} by performing all experiments on 4×3090 GPUs and training our agent for 400M steps. Following~\cite{al2022zero}, we set the navigation policy network as a 2-layer GRU with an embedding size of 128. 
We use a ResNet9 for feature extraction, and the size of the feature map in ``Dense Correlation'' and ``Direction-aware Correlation'' is 4×4 and 16×16, respectively. 
The layer number of the correlation pyramid in ``Direction-aware Correlation'' is set to 3~(maximum number of layers considering the input resolution). And we report the ImageNav results averaged over three random seeds.

\subsection{Comparison with State-of-the-art Methods}
In this section, we first compare the performance of our method with the state-of-the-art methods on the Gibson dataset. 
Then, following~\cite{al2022zero, sun2024fgprompt}, we directly test the Gibson-trained model on the unseen scenes from Matterport3D~(MP3D)~\cite{chang2017matterport3d} and HM3D~\cite{ramakrishnan2021habitat} for cross-domain evaluation to verify the ability of domain generalization. 

To enable the evaluation closer to the user experience, we add a new setting ``user-matched goal'' in addition to the original ``agent-matched goal'' setting~(as described in~\Cref{Setting}). We compare the results under these two settings for each of the three datasets.

\begin{table}[t]
\caption{Comparison of cross-domain generalization performance on MP3D. All methods are trained on Gibson and evaluated on MP3D. The two settings, ``agent-matched goal'' and ``user-matched goal'', represent the goal camera reflecting the task executor and task publisher, respectively. }
\label{tab:table2}
\centering
\resizebox{0.8\linewidth}{!}{
\begin{tabular}{lccc}
\hline
\textbf{Method}                      & \textbf{Backbone}    & \textbf{SPL}         & \textbf{SR}          \\\hline
\emph{\textbf{agent-matched goal}} & \multicolumn{1}{l}{} & \multicolumn{1}{l}{} & \multicolumn{1}{l}{} \\
NRNS~\cite{hahn2021no}                                 & ResNet18             & ~5.2\%                & ~9.3\%                \\
ZER~\cite{al2022zero}                                   & ResNet9              & 10.8\%               & 14.6\%               \\
FGPrompt~\cite{sun2024fgprompt}                              & ResNet9              & 44.9\%               & 75.5\%               \\
\rowcolor[HTML]{DAE8FC} 
RSRNav (ours)                       & ResNet9              & \textbf{51.7\%}      & \textbf{75.7\%}      \\\hline
\emph{\textbf{user-matched goal}}  & \multicolumn{1}{l}{} & \multicolumn{1}{l}{} & \multicolumn{1}{l}{} \\
ZER~\cite{al2022zero}                                   & ResNet9              & ~6.0\%                & ~6.7\%                \\
FGPrompt~\cite{sun2024fgprompt}                                     & ResNet9              & 27.9\%               & 55.3\%               \\
\rowcolor[HTML]{DAE8FC} 
RSRNav (ours)                       & ResNet9              & \textbf{39.8\%}      & \textbf{65.5\%}      \\\hline
\end{tabular}
}
\end{table}

\begin{figure*}[!t]
  \centering
  \includegraphics[width=0.8\linewidth]{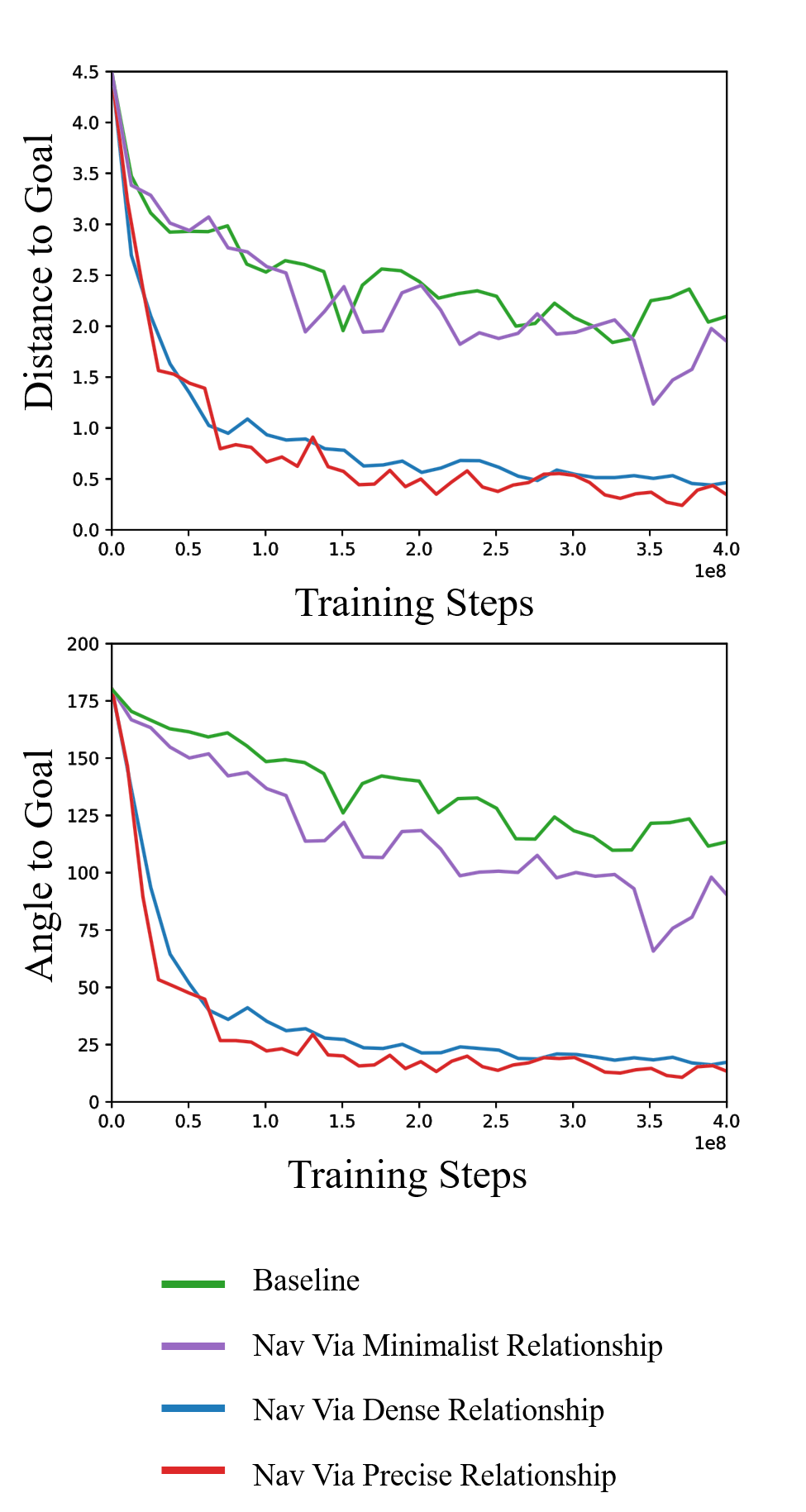}
  \caption{Comparison of convergence rates and tests results on metrics ``Distance to Goal'' and ``Angle to Goal''. The former metric measures the distance between the agent and the goal when stopping; the latter metric is the angle between the agent pose and the goal view when stopping. We also provide the average episode test results, plotted to the right of the convergence curve.}
  \label{fig:visual1}
\end{figure*}

\begin{table}[tb]
\caption{Comparison of cross-domain generalization performance on HM3D. All methods are trained on Gibson and evaluated on HM3D. All methods are trained on Gibson and evaluated on MP3D. The two settings, ``agent-matched goal'' and ``user-matched goal'', represent the goal camera reflecting the task executor and task publisher, respectively.}
\label{tab:table3}
\centering
\resizebox{0.8\linewidth}{!}{
\begin{tabular}{lccc}
\hline
\textbf{Method}                      & \textbf{Backbone}    & \textbf{SPL}         & \textbf{SR}          \\\hline
\emph{\textbf{agent-matched goal}} & \multicolumn{1}{l}{} & \multicolumn{1}{l}{} & \multicolumn{1}{l}{} \\
NRNS~\cite{hahn2021no}                                  & ResNet18             & ~4.3\%                & ~6.6\%                \\
ZER                                  & ResNet9              & ~6.3\%                & ~9.6\%                \\
FGPrompt~\cite{sun2024fgprompt}                               & ResNet9              & 39.2\%               & \textbf{74.2\%}      \\
\rowcolor[HTML]{DAE8FC} 
RSRNav (ours)                       & ResNet9              & \textbf{42.9\%}      & 72.9\%               \\
\hline
\emph{\textbf{user-matched goal}}  & \multicolumn{1}{l}{} & \multicolumn{1}{l}{} & \multicolumn{1}{l}{} \\
ZER~\cite{al2022zero}                                  & ResNet9              & ~3.9\%                & ~4.9\%                \\
FGPrompt~\cite{sun2024fgprompt}                             & ResNet9              & 24.2\%               & 52.9\%               \\
\rowcolor[HTML]{DAE8FC} 
RSRNav (ours)                       & ResNet9              & \textbf{30.1\%}      & \textbf{59.8\%}     \\\hline
\end{tabular}
}
\end{table}

\noindent{\bf Results on Gibson.}\hspace{0.3cm}
For the ``agent-matched goal'' setting, as shown in~\Cref{tab:table1}, our RSRNav outperforms the state-of-the-art methods, \textit{i.e.}, ranks first for metrics SPL, SR. Our baseline~(ZER) achieves 29.2\% success, with our RSRNav outperforming it 3-fold. Meanwhile, it achieves 21.6\% SPL, and our RSRNav outperforms it by 3.2x. The significant improvement demonstrates the effectiveness of our correlation-based method in ImageNav. 
Furthermore, our RSRNav shows considerable performance improvements compared to OVRL-V2, which uses a strong visual backbone~(ViT) with pre-training. Although employing a simpler backbone, ResNet9, without pre-training, we still perform better than it in both SPL~(69.5\% vs. 58.7\%) and SR~(91.1\% vs. 82.0\%). This suggests that correlation information is more critical for predicting navigation actions than the rich semantic features extracted by complex encoders.
Compared to Fgprompt, both using ResNet9, we surpass it by 3.0 for SPL and 0.7 for SR, the improvement in SPL demonstrates that our explicit modeling of correlation and the introduction of directional information can make navigation more efficient.

All methods are trained under ``agent-matched goal'' setting and then directly generalize to ``user-matched goal'' setting for evalution without any fine-tuning. This is challenging for the generalization and robustness of navigation methods. However, it also more emphatically showcases the superiority of our method under such a setting. Specifically, we substantially exceed FGPrompt by 9.8 for SPL and 7.2 for SR. This indicates that our correlation-based method is more robust and generalizable than previous methods.

\noindent{\bf Cross-Domain Generalization.}\hspace{0.3cm}
Beyond the visual domain disparity among these datasets, MP3D scenes are both more intricate and expansive compared to Gibson, while HM3D showcases a wide variety of scene types. This presents a very challenging cross-domain evaluation setting. Following the setting and testing episodes in ZER, we directly test our Gibson-trained model on these two new datasets without any tuning. As shown in~\Cref{tab:table2}, our RSRNav achieves the best results on SPL and SR under both two goal camera settings in MP3D. It is worth noting that our advantage is amplified under the ``user-matched goal'' setting, which again proves that our method is more generalizable and robust compared to previous methods. Similar results appear in HM3D, as shown in~\Cref{tab:table3}, although our success rate is lower than FGPrompt for setting ``agent-matched goal'' by 1.3, it is significantly higher in setting ``user-matched goal'' by 6.9.

\begin{table}[t]
\caption{Comparison of our three progressive correlation-based ImageNav methods on Gibson. Baseline* represents the feature map version baseline. All ablation studies are training 100 million steps using the Gibson dataset. ``MC'' represents the Minimalist Correlation. ``DC'' represents the Dense Correlation. ``Direct-C'' represens the Direction-aware Correlation. All methods are trained on Gibson and evaluated on MP3D. The two settings, ``agent-matched goal'' and ``user-matched goal'', represent the goal camera reflecting the task executor and task publisher, respectively. We report the ImageNav results averaged over three random seeds.}
\label{tab:table4}
\centering
\resizebox{1\linewidth}{!}{
\begin{tabular}{lcccc}
\hline
\multirow{2}{*}{\textbf{Setting}}          & \textit{\textbf{agent-matched}} & \textit{\textbf{goal}} & \textit{\textbf{user-matched}} & \textit{\textbf{goal}} \\
                                           & SPL                             & SR                     & SPL                            & SR                     \\ \hline
Baseline                                   & 13.4\%                          & 16.2\%                 & 12.7\%                         & 17.5\%                 \\
Baseline*                                  & 17.2\%                          & 20.0\%                 & 15.4\%                         & 18.2\%                 \\
Navigation Via MC      & 16.1\%                          & 22.6\%                 & 15.4\%                         & 21.6\%                 \\
Navigation Via DC           & 53.2\%                          & 79.4\%                 & 37.1\%                         & 64.0\%                 \\
Navigation Via Direct-C & 61.2\%                          & 86.2\%                 & 42.2\%                         & 69.2\%                 \\ \hline
\end{tabular}
}
\end{table}

\begin{table}[tb]
\caption{Ablation studies of ``Dense Correlation'' on Gibson.}
\label{tab:table5}
\centering
\resizebox{0.6\linewidth}{!}{
\begin{tabular}{cccc}
\hline
\textbf{\begin{tabular}[c]{@{}c@{}}Cross \\ Correlation\end{tabular}} & \textbf{\begin{tabular}[c]{@{}c@{}}Fine-grained \\ Correlation\end{tabular}} & \textbf{SPL} & \textbf{SR} \\ \hline
\ding{55}            & \ding{55}     & 16.1\%       & 22.6\%      \\
\checkmark               & \ding{55}     & 19.5\%       & 30.3\%      \\
\ding{55}              & \checkmark     & 36.5\%       & 53.6\%      \\
\checkmark               & \checkmark     & 53.2\%       & 79.4\%    \\ \hline
\end{tabular}
}
\end{table}

\begin{table}[tb]
\caption{Ablation studies of ``Direction-aware Correlation'' on Gibson.}
\label{tab:table6}
\centering
\resizebox{0.8\linewidth}{!}{
\begin{tabular}{ccccc}
\hline
\textbf{LookUp} & \textbf{\begin{tabular}[c]{@{}c@{}}Size of \\ Feature Map\end{tabular}} & \textbf{\begin{tabular}[c]{@{}c@{}}Correlation Pyramid\\ (number of layers)\end{tabular}} & \textbf{SPL} & \textbf{SR} \\ \hline
\ding{55}     & 4×4              & 1                                                                                         & 53.2\%       & 79.4\%      \\
\checkmark      & 4×4             & 1                                                                                         & 50.4\%       & 75.4\%      \\
\checkmark      & 16×16             & 1                                                                                         & 58.0\%       & 83.7\%      \\
\checkmark      & 16×16             & 2                                                                                         & 59.4\%       & 86.3\%      \\
\checkmark      & 16×16             & 3                                                                                         & 61.2\%       & 86.2\%      \\ \hline
\end{tabular}
}
\end{table}

\subsection{Ablation Study}
In this section, we report and analyze the performance of the baseline and three progressive versions of our proposed correlation-based method. Then we analyze the evolution of our correlation-based method from ``Minimalist Correlation'' to ``Dense Correlation'' and ``Dense Correlation'' to ``Direction-aware Correlation'' one by one. All ablation studies are training 100 million steps using the Gibson dataset.

\noindent{\bf Three Progressive Correlation-based Method.}\hspace{0.3cm}
First, we establish the baseline setting, we follow ZER~\cite{al2022zero} to construct our Baseline, which is the vector version as same as ZER. For a more fair and complete comparison of our methods, we constructed the feature map version and formed Baseline*. Baseline* is based on Baseline, and the concatenation of the feature vector is replaced with the concatenation of the feature map passed to the policy network, retaining more spatial information.

The ``Navigation Via Minimalist Correlation'' version is designed to explore the importance of correlation among a large pile of information, which only passes two correlation scores to the policy network. As illustrated in~\Cref{tab:table4}, it outperforms Baseline by~(+2.7 SPL, +6.4 SR), indicating that despite the low information content of ``Minimalist Correlation'', it is more useful and effective and that there is information redundancy in Baseline. With this exciting phenomenon, we further enhance the construction and utilization of correlation and evolve it to the ``Navigation Via Dense Correlation'' version. Both metrics improved by more than 3x with the richer amount of correlation information, further confirming the critical role of correlation in ImageNav. Also, with the same processing of the feature maps we bring 4× improvements in the success rate and 3× improvements in SPL comparing with Baseline*. Finally, we introduce ``Direction-aware Correlation'' to provide more direction information to cue ImageNav, which brings a great boost~(+8.0 SPL, +6.8 SR).

\noindent{\bf Analysis of Dense Correlation.}\hspace{0.3cm} 
Evolution from ``Minimalist Correlation'' to ``Dense Correlation'', there are two key points in the construction of dense correlation: cross-correlation and fine-grained correlation. Here, we explore the contribution of the two key points to performance improvement. As shown in~\Cref{tab:table5}, both can bring performance gains, fine-grained correlation significantly improves information content while cross correlation allows the global relation between goal and egocentric images. Although they can improve performance separately, together, they can maximize benefits. Essentially, both are increasing the amount of correlation information, demonstrating that as the correlation information gets richer, the performance gets better.

\begin{figure}[tb]
  \centering
  \includegraphics[width=0.6\linewidth]{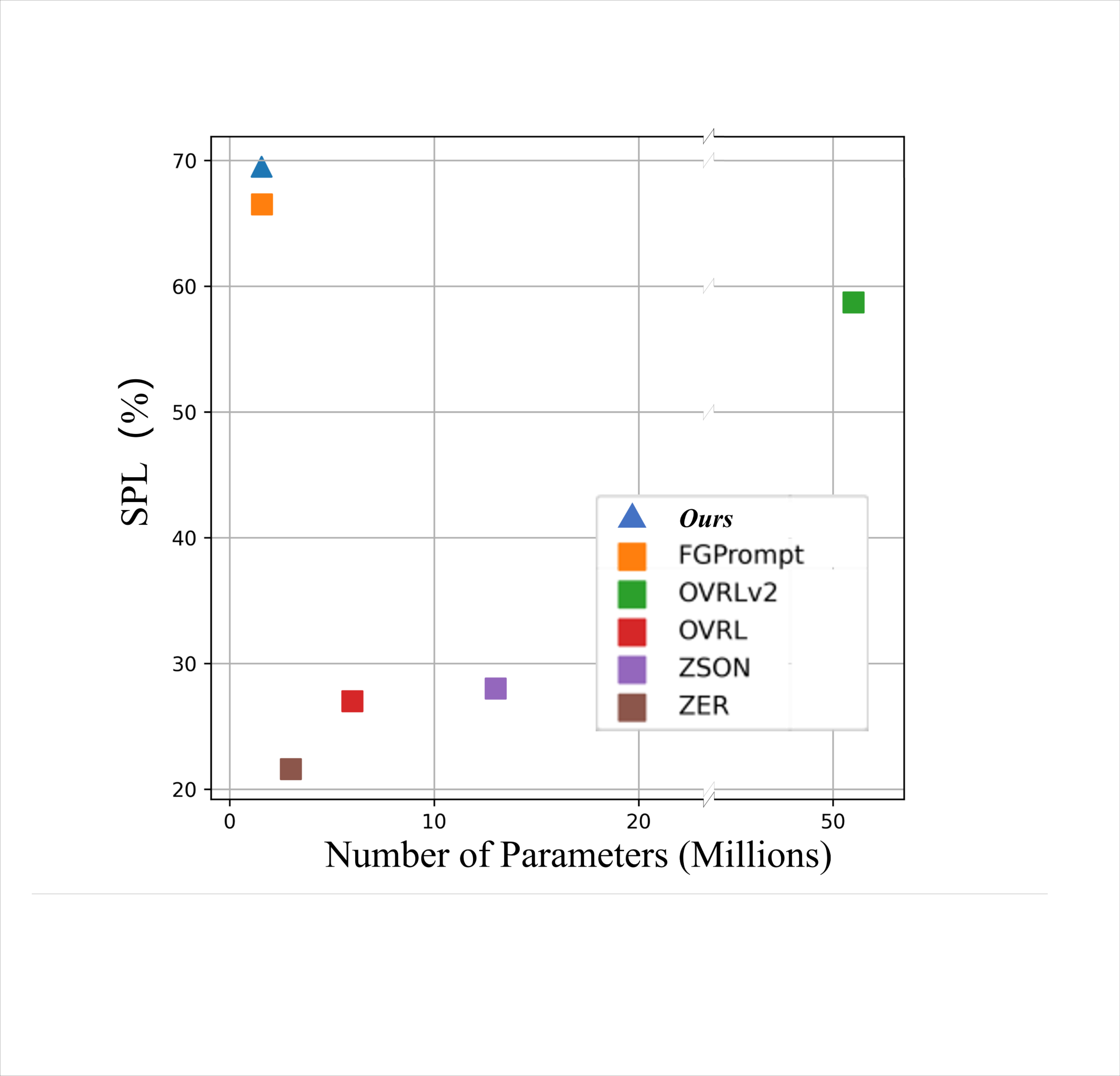}
  \caption{Compare the number of parameters, as well as the SPL on Gibson.}
  \label{fig:para}
\end{figure}

\noindent{\bf Analysis of Direction-aware Correlation.}\hspace{0.3cm} 
The first and last rows of~\Cref{tab:table6} present the performance with ``Dense Correlation'' and ``Direction-aware Correlation'', respectively. The results of the first three lines show that when we simply replace cross-correlation with the LookUp operation, it brings about performance degradation, which only works if we increase the size of the feature map. We then conduct experiments on the correlation pyramid as shown in the last three rows of~\Cref{tab:table6}. Shows that as the number of layers increases, the navigation performance improves. As seen in~\Cref{fig:method2}, the multi-layer pyramid yields a multi-scale directional correlation; the more layers there are, the richer the correlation information is. These demonstrate the validity of our direction-aware correlation, and a more qualitative analysis of direction information is in the visualizations below.

\begin{figure*}[!t]
  \centering
  \includegraphics[width=1\linewidth]{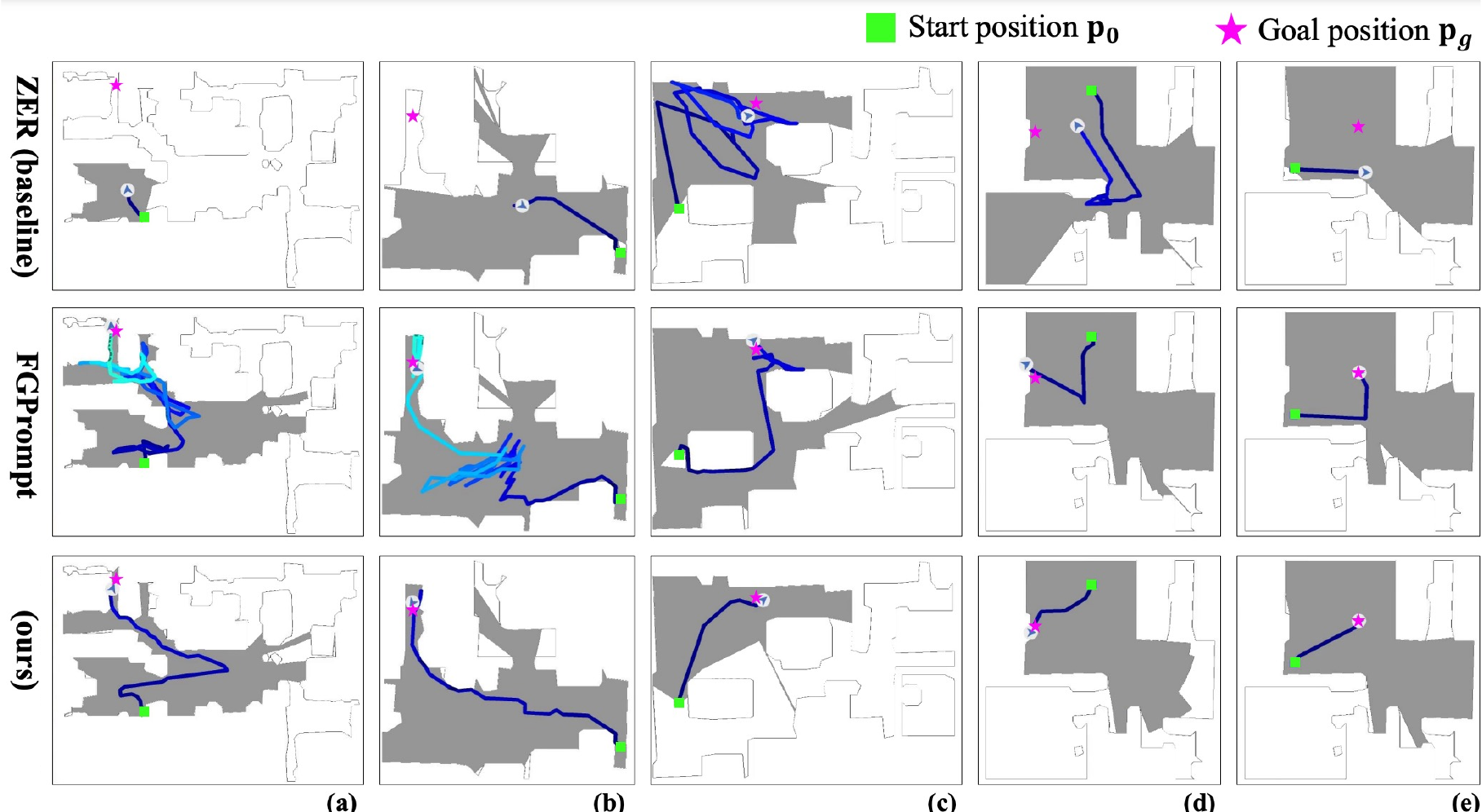}
  \caption{Comparison of Visualization Results. The gray area in the scene represents the area scanned by the egocentric camera while the agent is traveling. The agent's navigational path is indicated by a blue line that gets brighter as the time steps get longer~(from dark blue to light blue). These are the results under the ``user-matched camera'' setting.}
  \label{fig:visual2}
\end{figure*}

\begin{figure*}[!t]
  \centering
  \includegraphics[width=1\linewidth]{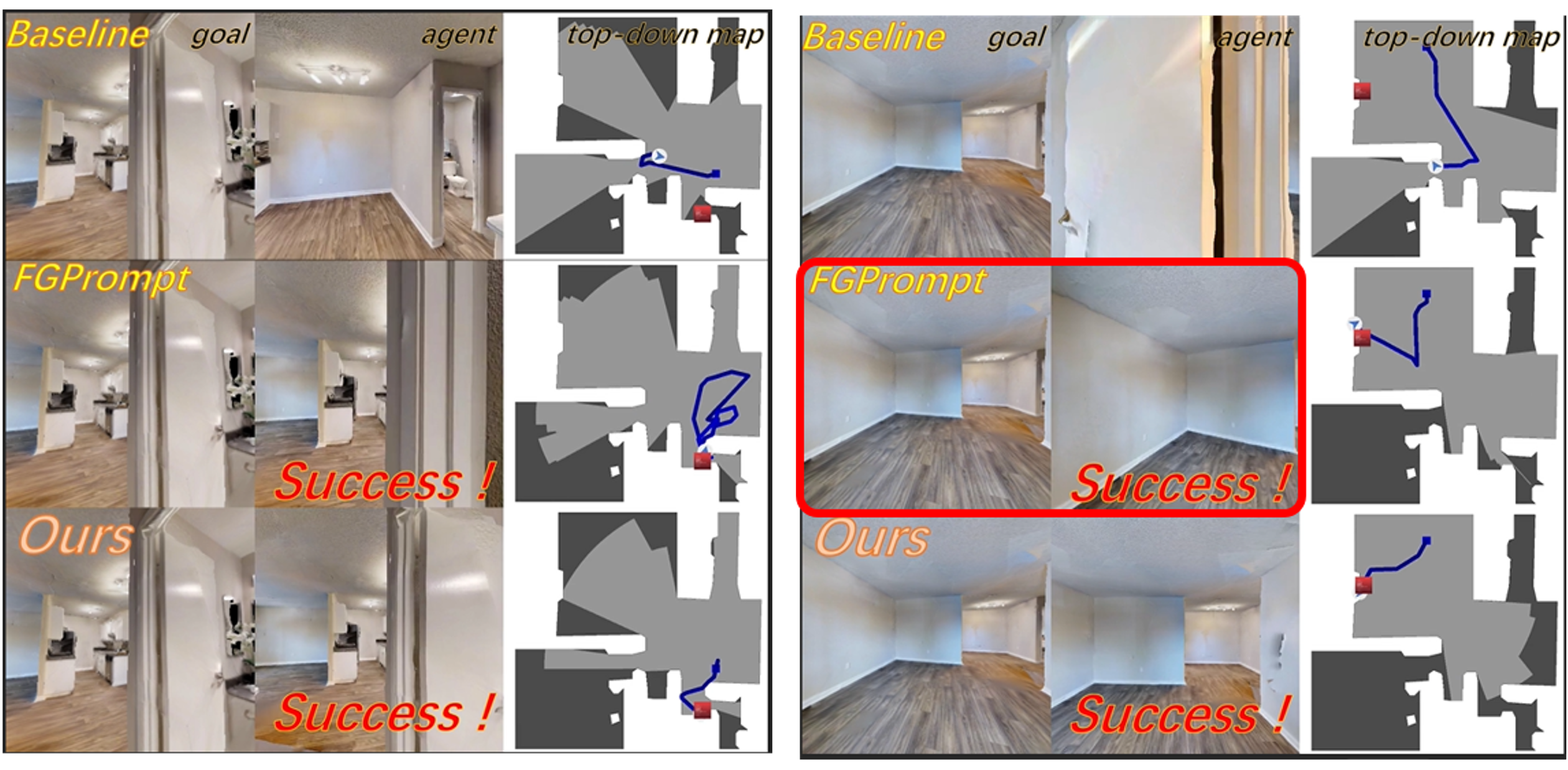}
  \caption{Comparison of three methods. The left side shows the target viewpoint image, the middle shows the image captured by the agent's own camera after reaching the target location, and the right side shows the overhead map. As can be seen from the red boxes, although FGPrompt found the target location and was marked as successful, there is a deviation in accuracy.}
  \label{fig:visual3}
\end{figure*}

\noindent{\bf Performance vs. Model Size.}\hspace{0.3cm} 
Smaller AI models bring significant benefits to real-world deployments by requiring less computational power, enabling them to run on a wider range of devices with less energy consumption. This is even more important for an embodied intelligence that requires visual navigation, as it has to carry other operational facilities in addition to what it needs to navigate, so the computational resources required for navigation are stringent. As shown in~\Cref{fig:para}, we compare the number of parameters, as well as the SPL on Gibson for~\cite{al2022zero, sun2024fgprompt, yadav2023offline, yadav2023ovrl, majumdar2022zson}. Our RSRNav outperforms existing methods and has the smallest number of parameters among all methods, showing potential for real-world applications.

\noindent{\bf More Extreme ``user-matched goal'' Setting.}\hspace{0.3cm} 
When used in the real world, there should not be too much restriction on the goal camera as the user can shoot the goal image at any height and from any angle. In addition to the two goal settings in the main text, \emph{i.e.} ``agent-matched goal'' and ``user-matched goal'', we add a more extreme setting of goal camera in this part. Specifically, we sample the height of the goal camera from $\mathcal{U}(0.8m, 1.5m)$, the pitch delta from $\mathcal{U}(-45^\circ, 45^\circ)$, and the HFOV from $\mathcal{U}(60^\circ, 120^\circ)$. The results reported in~\Cref{tab:table7} show that our method significantly surpasses both the baseline and state-of-the-art method FGPropm in performance.

\begin{table}[t]
\caption{Additional comparison under ``extreme setting'' on Gibson.}
\label{tab:table7}
\centering
\resizebox{1\linewidth}{!}{
\begin{tabular}{l|cc|cc|cc}
\hline
\multirow{2}{*}{\textbf{Method}} & \multicolumn{2}{c|}{\textbf{agent-matched goal}} & \multicolumn{2}{c|}{\textbf{user-matched goal}} & \multicolumn{2}{c}{\textbf{extreme setting}} \\
                                 & SPL                     & SR                     & SPL                    & SR                     & SPL                   & SR                   \\ \hline
ZER                              & 21.6\%                  & 29.2\%                 & 12.8\%                 & 15.7\%                 & 11.3\%                & 13.5\%               \\
FGPrompt                         & 66.5\%                  & 90.4\%                 & 46.8\%                 & 76.0\%                 & 16.1\%                & 31.2\%               \\
RSRNav (ours)                   & \textbf{69.5\%}         & \textbf{91.1\%}        & \textbf{56.6\%}        & \textbf{83.2\%}        & \textbf{22.8\%}       & \textbf{40.2\%}      \\ \hline
\end{tabular}
}
\end{table}

\subsection{Analysis of Correlation-based method for ImageNav}
We analyze human-intuitive navigation, and inspired by it we identify two limitations of previous methods: (1) Semantic feature vectors inadequately convey valid direction information for navigation, resulting in inefficient and superfluous actions; (2) Performance decreases dramatically when encountering viewpoint inconsistencies caused by differences in camera settings between training and application. FGPrompt has not explicitly modeled the comparison between the current observation to the goal. They only provide the feature vectors, putting the full burden of comparison on the policy network and complicating the training. Semantic feature vectors lack precise spatial details, resulting in unclear directions for the policy and potentially causing unnecessary actions during navigation towards the goal. 
In this part, we analyze how our proposed correlation-based method addresses these challenges. For the first challenge, we analyze and experimentally verify that the correlation is the key information to guide navigation actions. Therefore, instead of directly passing the features in the perception step, we construct the correlation between them and then allow the policy network to predict action based on it. This can significantly reduce the learning burden of the policy network. As shown in~\Cref{fig:visual1}, ``Navigation Via Dense Correlation'' and ``Navigation Via Direction-aware Correlation'' converge significantly faster with sufficient correlation information given to the policy network. For the second challenge, whether the angle can be adjusted accurately according to the goal can be reflected most intuitively in the ``Angle to Goal'' metric, which is the angle between the agent pose and the goal view when stopping. Comparison of convergence curves and test results for this metric, as shown in~\Cref{fig:visual1}, demonstrates that correlation can help the agent adjust the pose accurately. Then, after we introduce direction-aware correlation, the precision of the agent's pose becomes higher.

\subsection{Visualization}
We visualize the navigation results under the ``user-matched goal'' setting, with the navigation path visualized on the top-down maps. We compare our RSRNav with the other two methods for different levels of difficulty in navigation in~\Cref{fig:visual2}~(from right to left, navigation goes from easy to hard). 

For the hard situations (a) and (b), ZER fails the navigation, and FGPrompt succeeds but takes a lot of detours. In contrast, our RSRNav efficiently explores new environments when no goal-relevant view is seen, approaches the goal as soon as the agent sees it, and stops in time when reaching.
For the medium situations (c), FGPrompt doesn't stop in time when passing the goal, and we give the stop action the first time we reach the goal. For easy cases (d) and (e), there is no need to explore a new environment, and the goal view can be seen at the initial location. In contrast to the other two methods, our RSRNav localizes directly to the goal and takes an almost straight path to reach it. This is since the correlation between the current observation and the goal can guide policy network with precise directional information.

\noindent{\bf Comparison of viewpoints of the target position.}\hspace{0.3cm} 
We provide video demos in the zip file, showcasing a comparison of the agent performance across our RSRNav, the Baseline, and the state-of-the-art method FGPropmt within identical navigation goals and scenarios. The gray area in the scene represents the area scanned by the egocentric camera while the agent is traveling. The agent’s navigational path is indicated by a blue line that gets brighter as the time steps get longer (from dark blue to light blue). The ``\emph{Succees~!}'' symbol appears when the target successfully completes the navigation task. These are the results under the ``user-matched goal'' setting. 

We intercepted two of these navigation result comparisons as shown in \Cref{fig:visual3}. The success of a navigation task is judged by the distance to the target after the “stop” command. However, as shown in the red box in the figure, even though the previous method can successfully navigate to the target location, it is still not precise enough for the angle of view, whereas we can make the navigation angle more precise, thanks to the modeling of the orientation information.

\section{Conclusion}\label{sec:conclusion}
We propose a novel method for efficient and robust ImageNav by reasoning spatial relationship between the goal image with current observations during the navigation process. In detail, we construct correlations between the goal and current observations to explicitly model the spatial relationship and pass them to the policy network for action prediction. And we analyze and experimentally reveal that the spatial relationship plays a the key role in navigation action prediction. Moreover, we gradually enhance the relationship modeling by using more fine-grained cross-correlation and introducing direction-aware correlation to provide precise direction information for navigation.
Extensive evaluation of our RSRNav on 3 benchmark datasets~(Gibson, HM3D, and MP3D) shows superior performance in terms of navigation efficiency~(SPL). Furthermore, RSRNav significantly outperforms previous state-of-the-art methods across all metrics under the “user-matched goal” setting. 
The results shows that our method has excellent performance and strong generalization capabilities to unseen scenes and inconsistent ``user-matched goal'' setting. We hope our RSRNav can serve as a baseline of relationship-based ImageNav method for future research.

\noindent{\bf Limitation and Future Work.}  
All the data for training our RSRNav is sourced from the simulator, which has gaps from the real-world scene. 
We plan to drive our RSRNav to navigate in the real-world by pre-training it on real-world correlation related datasets.



\ifCLASSOPTIONcaptionsoff
  \newpage
\fi

\bibliographystyle{IEEEtran}
\bibliography{IEEEabrv,referemce}

\begin{IEEEbiography}
[{\includegraphics[width=1in,height=1.25in,clip,keepaspectratio]{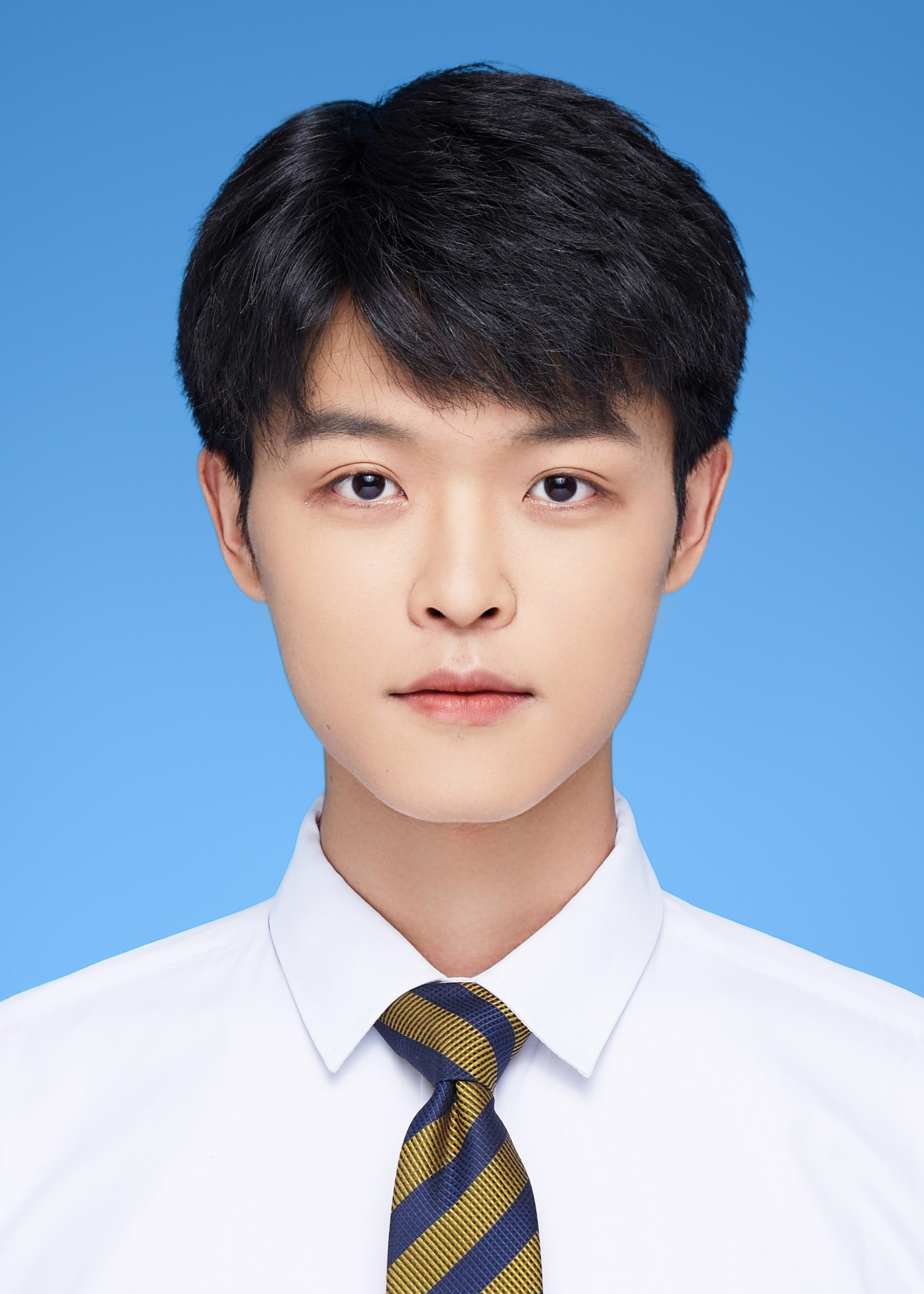}}]{Zheng Qin}
received the B.S. degree in robotic engineering from the Harbin Institute of Technology, China, in 2021. He is currently working toward the PdD. degree in artificial intelligence from Xi'an Jiaotong University. His research interests include multi-object tracking, visual navigation, motion generation and video generation.
\end{IEEEbiography}

\begin{IEEEbiography}[{\includegraphics[width=1in,height=1.25in,clip,keepaspectratio]{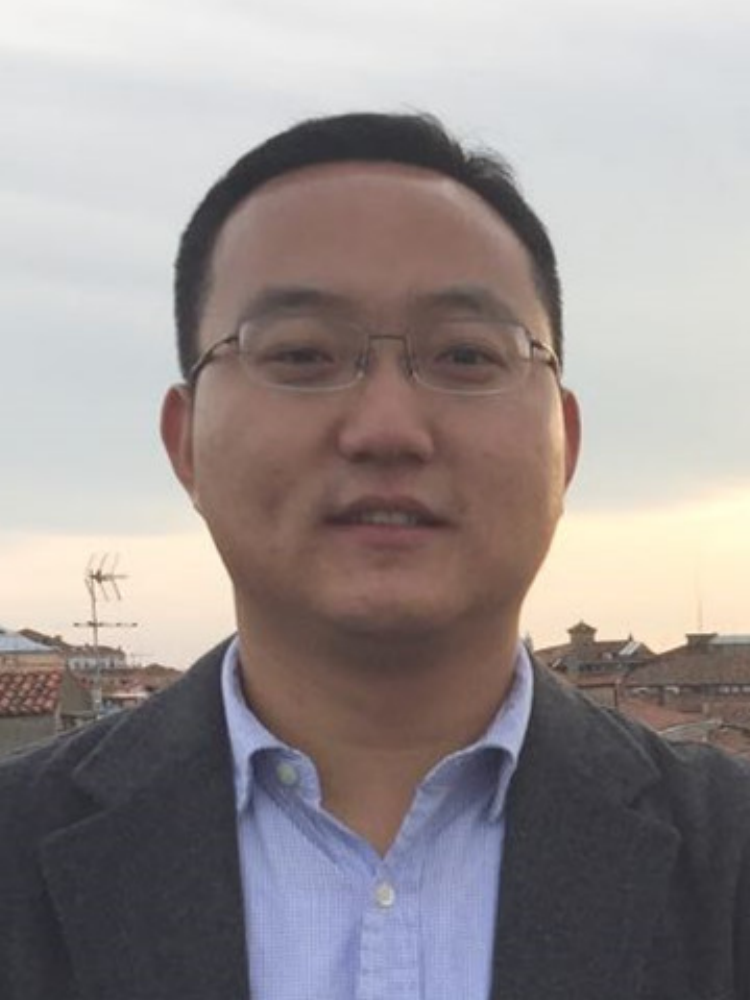}}]{Le Wang}(Senior Member, IEEE) received the B.S. and Ph.D. degrees in Control Science and Engineering from Xi'an Jiaotong University, Xi'an, China, in 2008 and 2014, respectively. From 2013 to 2014, he was a visiting Ph.D. student with Stevens Institute of Technology, Hoboken, New Jersey, USA. From 2016 to 2017, he is a visiting scholar with Northwestern University, Evanston, Illinois, USA. He is currently a Professor with the Institute of Artificial Intelligence and Robotics of Xi'an Jiaotong University, Xi'an, China. His research interests include computer vision, pattern recognition, and machine learning.
\end{IEEEbiography}

\begin{IEEEbiography}
[{\includegraphics[width=1in,height=1.25in,clip,keepaspectratio]{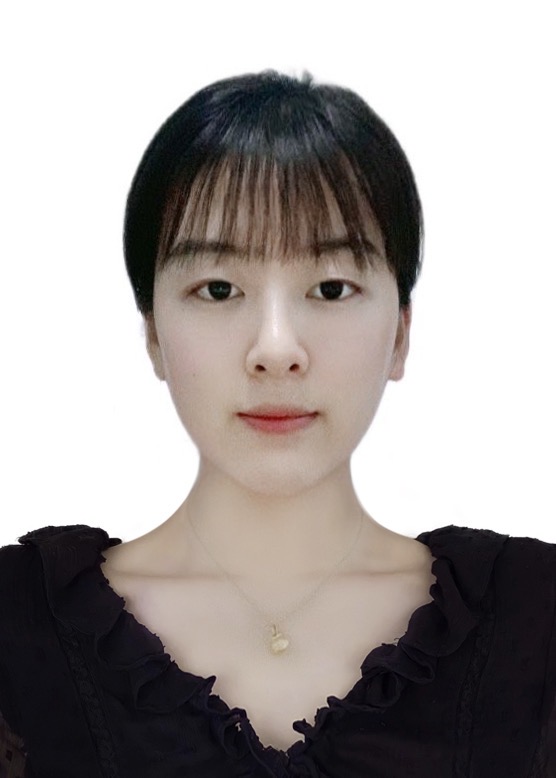}}]{Yabing Wang}
received the BE degree in software engineering from Zhengzhou University, Zhengzhou, China, in 2020, and the ME degree in College of Computer Science and Technology from Zhejiang Gongshang University, Hangzhou, China, in 2023. She is currently working toward the PhD degree with the Institute of Artificial Intelligence and Robotics, Xi’an Jiaotong University. Her research interests include multi-modal learning.
\end{IEEEbiography}

\begin{IEEEbiography}[{\includegraphics[width=1in,height=1.25in,clip,keepaspectratio]{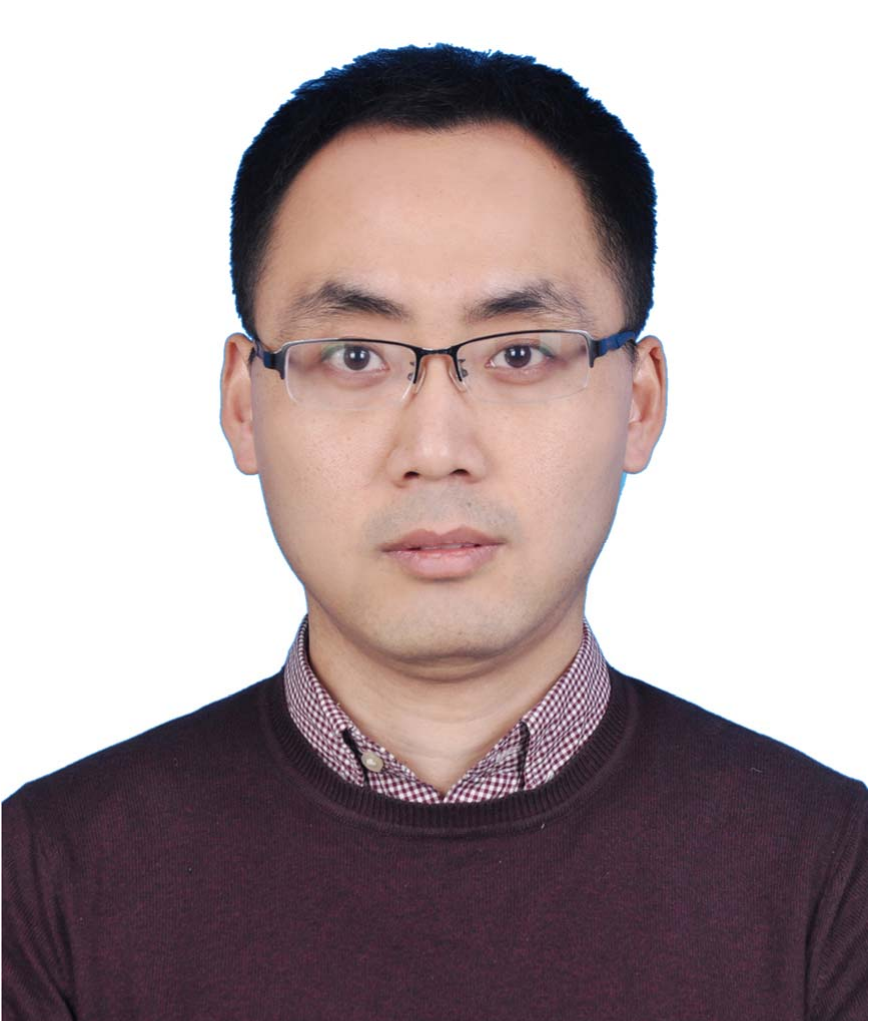}}]{Sanping Zhou} (Member, IEEE)
	received Ph.D. degree in control science and engineering from Xian Jiaotong University, Xi'an, China, in 2020. From 2018 to 2019, he was a visiting Ph.D. student at the Robotics Institute, Carnegie Mellon University. He is currently an Associate Professor with the Institute of Artificial Intelligence and Robotics, Xian Jiaotong University, Xi'an, China. His research interests include machine learning and computer vision, with a focus on object detection, image segmentation, visual tracking, multitask learning, and metalearning.
\end{IEEEbiography}

\begin{IEEEbiography}[{\includegraphics[width=1in,height=1.25in,clip,keepaspectratio]{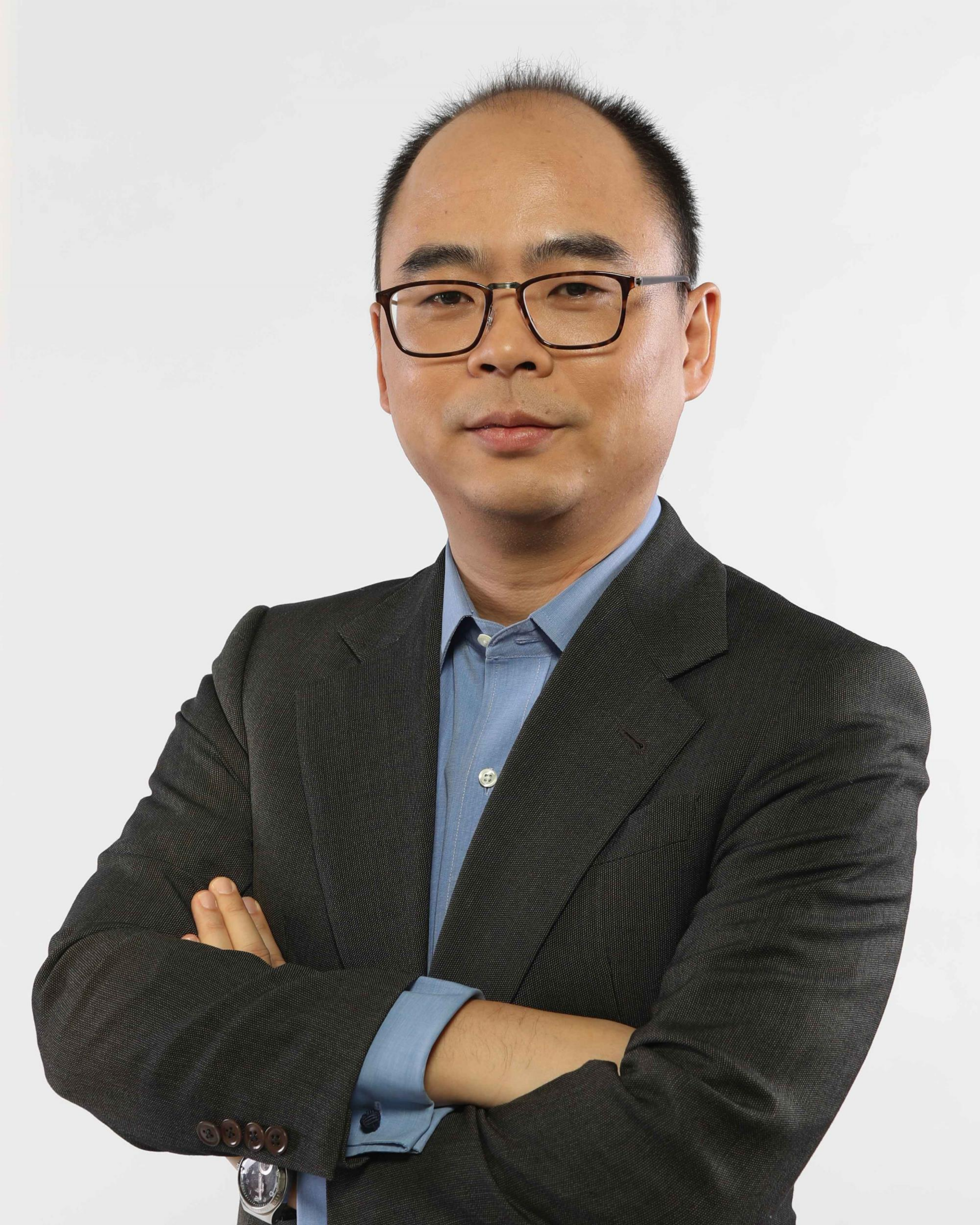}}]{Gang Hua} (Fellow, IEEE) received the B.S. and M.S. degrees in Automatic Control Engineering from Xi'an Jiaotong University (XJTU), Xi'an, China, in 1999 and 2002, respectively. He received the Ph.D. degree in Electrical Engineering and Computer Science at Northwestern University, Evanston, Illinois, USA, in 2006. He is currently the Vice President of the Multimodal Experiences Research Lab at Dolby Laboratories. His research focuses on computer vision, pattern recognition, machine learning, robotics, towards general Artificial Intelligence, with primary applications in cloud and edge intelligence. Before that, he was the CTO of Convenience Bee, and  the Managing Director and Chief Scientist of its research branch in US, Wormpex AI Research (2018-2024). He also served in various roles at Microsoft (2015-18) as the Science/Technical Adviser to the CVP of the Computer Vision Group, Director of Computer Vision Science Team in Redmond and Taipei ATL, and Senior Principal Researcher/Research Manager at Microsoft Research . He was an Associate Professor at Stevens Institute of Technology (2011-15). During 2014-15, he took an on leave and worked on the Amazon-Go project. He was a Visiting Researcher (2011-14) and a Research Staff Member (2010-11) at IBM Research T. J. Watson Center, a Senior Researcher (2009-10) at Nokia Research Center Hollywood, and a Senior Scientist (2006-09) at Microsoft Live labs Research. He is an associate editor of TPAMI and MVA. He is a general chair of ICCV'2027 and a program chair of CVPR'2019\&2022. He is the author of more than 200 peer reviewed publications in prestigious international journals and conferences. He holds 35 US patents and has 15 more US patents pending. He is the recipient of the 2015 IAPR Young Biometrics Investigator Award. He is an IEEE Fellow, an IAPR Fellow, and an ACM Distinguished Scientist.
\end{IEEEbiography}

\begin{IEEEbiography}[{\includegraphics[width=1in,height=1.25in,clip,keepaspectratio]{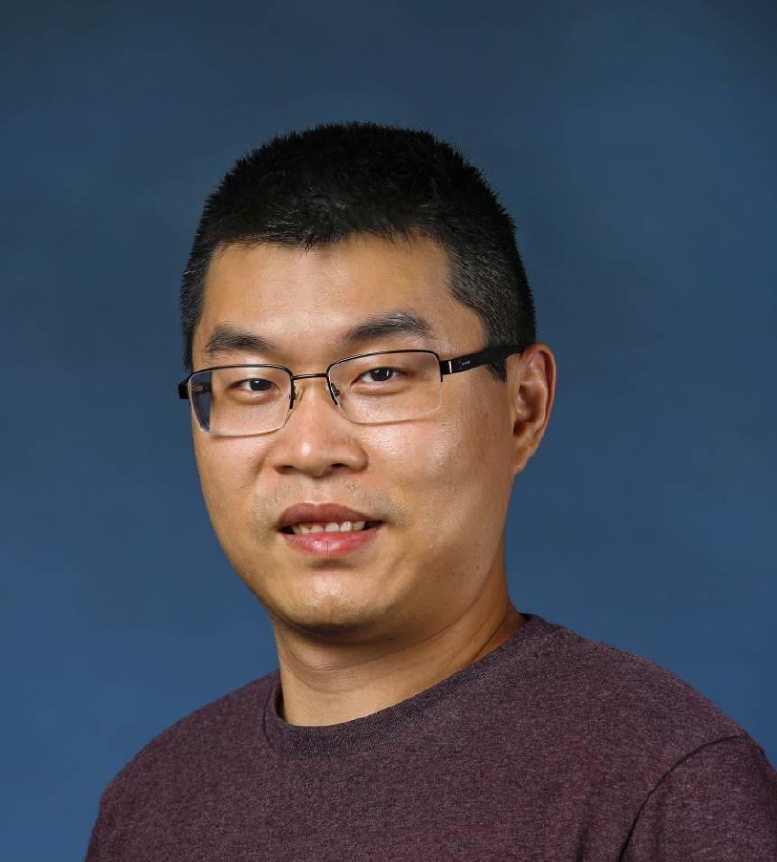}}]{Wei Tang} (Member, IEEE)
	received his Ph.D. degree in Electrical Engineering from Northwestern University, Evanston, Illinois, USA in 2019. He received the B.E. and M.E. degrees from Beihang University, Beijing, China, in 2012 and 2015 respectively. He is currently an Assistant Professor in the Department of Computer Science at the University of Illinois at Chicago. His research interests include computer vision, pattern recognition and machine learning.

\end{IEEEbiography}




\end{document}